\documentclass[a4papper,review,sort&compress,preprint,12pt]{elsarticle}

\usepackage{amssymb}
\usepackage{amsmath}
\usepackage{booktabs}
\usepackage{multirow}
\usepackage{color, xcolor}
\usepackage{array}

\journal{Neurocomputing}

\usepackage{caption}
\begin{document}

\begin{frontmatter}



\title{A Dual Domain Multi-exposure Image Fusion Network based on the Spatial-Frequency Integration}            

%
\author[1]{Guang Yang}
\ead{gyang2014@stu.xidian.edu.cn}


\author[1]{Jie Li}
\ead{leejie@mail.xidian.edu.cn}

\author[1,2]{Xinbo Gao \corref{cor1}}
\ead{xbgao@mail.xidian.edu.cn}

%
\cortext[cor1]{Corresponding author}

\affiliation[1]{organization={State Key Laboratory of Integrated Services Networks, School of Electronic Engineering, Xidian University},
    city={Xi'an},
    postcode={Shaanxi 710071}, 
    country={China}}

\affiliation[2]{organization={Chongqing Key Laboratory of Image Cognition, Chongqing University of Posts and Telecommunications},
    city={Chongqing},
    postcode={Chongqing 400065}, 
    country={China}}

\begin{abstract}
Multi-exposure image fusion aims to generate a single high-dynamic image by integrating images with different exposures. Existing deep learning-based multi-exposure image fusion methods primarily focus on spatial domain fusion, neglecting the global modeling ability of the frequency domain. To effectively leverage the global illumination modeling ability of the frequency domain, we propose a novelty perspective on multi-exposure image fusion via the Spatial-Frequency Integration Framework, named MEF-SFI. Initially, we revisit the properties of the Fourier transform on the 2D image, and verify the feasibility of multi-exposure image fusion on the frequency domain where the amplitude and phase component is able to guide the integration of the illumination information. Subsequently, we present the deep Fourier-based multi-exposure image fusion framework, which consists of a spatial path and frequency path for local and global modeling separately. Specifically, we introduce a Spatial-Frequency Fusion Block to facilitate efficient interaction between dual domains and capture complementary information from input images with different exposures. Finally, we combine a dual domain loss function to ensure the retention of complementary information in both the spatial and frequency domains. Extensive experiments on the PQA-MEF dataset demonstrate that our method achieves visual-appealing fusion results against state-of-the-art multi-exposure image fusion approaches. Our code is available at https://github.com/SSyangguang/MEF-freq.
\end{abstract}


\end{frontmatter}


\section{Introduction}

Illumination in the natural scene usually exhibits significant variations. For instance, during the daytime, the brightness can range from $10^5~cd/m^2$ (sunlight) to $10^2~cd/m^2$ (room light), and can even be reduced to $10^{-3}~cd/m^2$ at night \cite{xu2022multi}. Since the technical limitations of the imaging devices, a single low dynamic range (LDR) image captured by the sensoris insufficient to fully preserve content information in both over-exposed and under-exposed regions of the natural scene, resulting in unsatisfactory visual effects like structure distortions and poor color saturation \cite{karim2023current, nayar2000high}. Image fusion is designed to combine essential information from multiple source images and create a single image that is informative for human perception \cite{li2017pixel}, such as multi-modality image fusion \cite{li2018densefuse, song2022triple, zhao2023cddfuse}, digital photography image fusion \cite{liu2021multiscale, liu2022multi} and sharping fusion \cite{yang2017pannet, xu2020sdpnet}. Multi-exposure image fusion (MEF) is one kind of digital photography image fusion that can combine complementary information of images with different exposure settings, and generate a high dynamic range (HDR) image with visually appealing detail information both in over-/under-exposed areas. The key issue of MEF is to ensure the consistency of detail information from the over-/under- exposed regions in the fusion results \cite{liu2023holoco}.

Traditional MEF methods can be divided into two categories: spatial domain-based and transform domain-based, which fuse images via the hand-crafted features and transformation separately \cite{zhang2021benchmarking}. Kinoshita et al. \cite{kinoshita2019scene} proposed a scene segmentation-based method to adjust luminance and generate high-quality images. Ma et al. \cite{ma2017robust} proposed a structural patch decomposition method named SPD-MEF. Kong et al. \cite{kong2022guided} proposed a guided filter random walk and improved spiking cortical model-based general image fusion method in the NSST domain. Li et al. \cite{li2013image} adopted the guided filtering-based weighted average approach for general image fusion. Recently, deep learning (DL)-based MEF approaches have been applied to generate visually pleasing images and achieve great progress. Kalantari et al. \cite{ram2017deepfuse} was the first attempt to implement MEF via the convolutional network, and optimized the network in a supervised manner with MEF-SSIM \cite{ma2015perceptual}. Researchers \cite{ma2019deep, xu2020u2fusion, zhang2020ifcnn} proposed typical fully convolutional neural network (CNN) methods to combine images with varying exposure levels. Xu et al. \cite{xu2020mef} and Yang et al. \cite{yang2021ganfuse} proposed representative generative adversarial network (GAN)-based methods that incorporated adversarial game into the network training to implicitly fulfill feature extraction, feature fusion and image reconstruction. Liu et al. \cite{liu2023holoco} introduced contrastive learning to MEF, representing the source LDR image, reference HDR image and the fusion HDR image as the negative, positive and anchor, respectively. Xiang et al. \cite{xiang2022recognition} applied multi-exposure image fusion in industry online scene text recognition, specifically by synthesizing detail-enhanced images from multi-exposure image sequences to increase the contrast between characters and backgrounds. However, above mentioned DL-based methods only extracted and integrated complementary information in the spatial domain, and resulted in exposure inconsistency due to the local property of the convolutional operator. Spatial and frequency interaction was introduced in pan-sharpening by \cite{zhou2022spatial}, while exposure correction in the frequency domain was addressed by \cite{huang2022deep, li2022embedding}. Yu et al. \cite{yu2022frequency} explored the frequency and spatial dual guidance in image dehazing. Qu et al. \cite{qu2023aim, qu2023rethinking} exploited frequency information to tackle two important challenges, which were mining hard-to-learn features and synthesizing images with diverse exposure levels methods through the Fourier transformation. However, a multi-exposure image fusion framework that incorporates joint spatial and frequency information interaction is still under-exploited.

\begin{figure}[!t]
\centering
\includegraphics[width=5.5in]{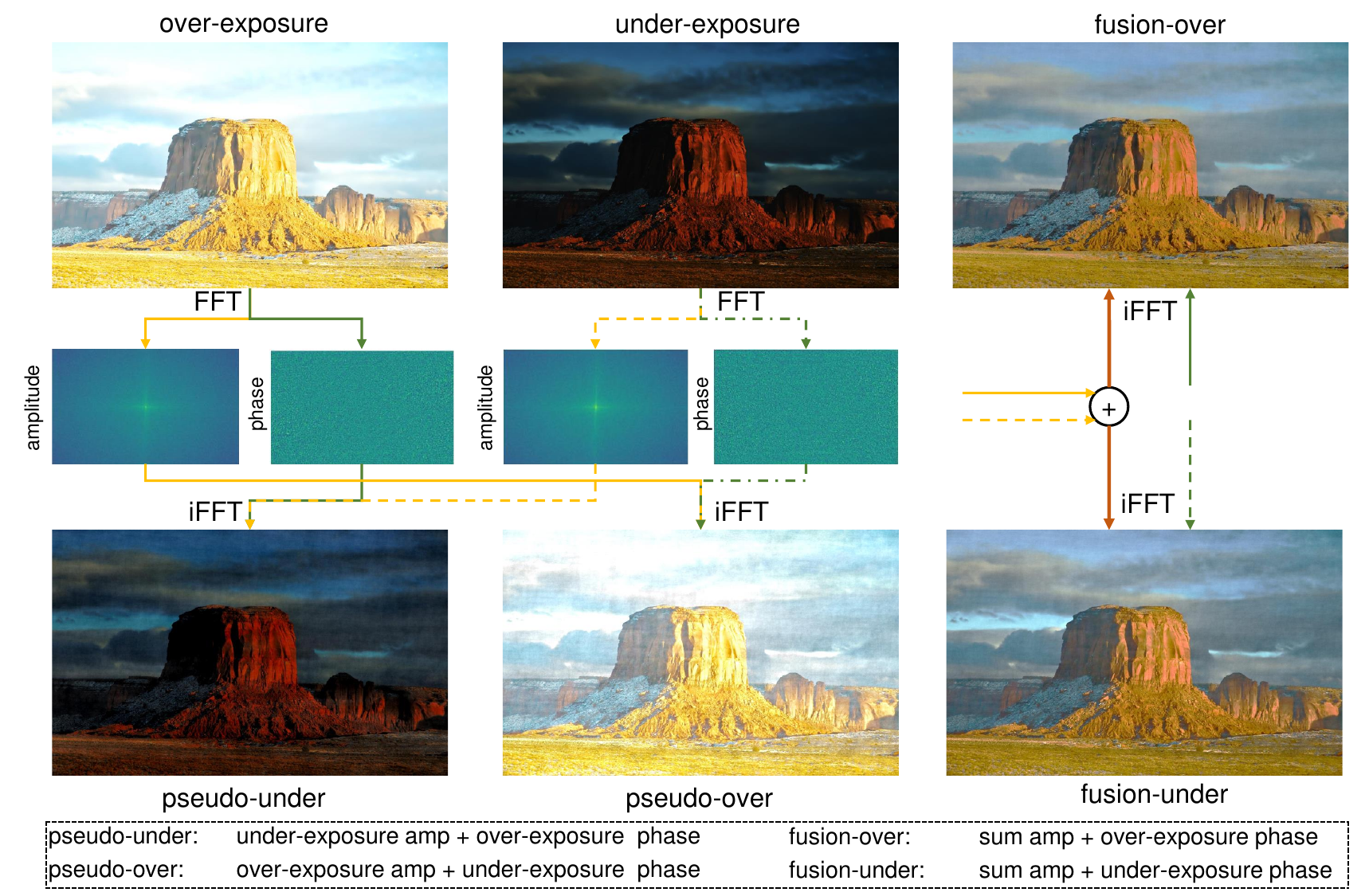}
\caption{
Visualization results of reconstructed images of swapping amplitude and phase components. The left side images are reconstructed images of swapping the amplitude and phase components. The right side images are reconstructed images of fused amplitude components and phase components of the original image pairs.
}
\label{fig:analysis}
\end{figure}

In \cite{oppenheim1981importance, skarbnik2009importance}, the amplitude component can represent luminance information and the phase component conveys structure information in the case of 2D images, as illustrated in Fig. \ref{fig:analysis}. In Fig. \ref{fig:analysis}, we visualize the function of the amplitude component in representing illumination by converting the over-exposed and under-exposed images to the Fourier space and exchanging their respective amplitude components while preserving the original phase components. To further explore the potential ability of the frequency domain in acquiring visual-friendly multi-exposure image fusion results, we conducted additional experiments to validate the possibility of integrating illumination information through the frequency domain, i.e.utilizing weighted sum to combine amplitude component and then converting back to the spatial domain, as shown in the right side of the Fig. \ref{fig:analysis}.

From the Fig. \ref{fig:analysis}, we can draw the following two conclusions. Firstly, the visualization result obtained by combining the amplitude of the over-exposed image with the phase of the under-exposed image is consistent with the brightness of the original over-exposed image, and so is the under-exposed one. Secondly, the brightness of the amplitude-weighted sum result ranges between the brightness exhibited by the over-/under-exposed images. This property makes it suitable for multi-exposure image fusion. Given that images with different exposures in the same scene share the same semantic representations and different brightness levels in images \cite{land1977retinex}, it is natural to fuse images in the frequency domain, which is able to be considered as the global complement to the spatial domain fusion to solve the exposure inconsistency.

Based on the aforementioned analysis, we propose the Spatial-Frequency Integration Framework (MEF-SFI) to introduce an image-wide receptive field over the frequency domain to enhance complementary information integration in the MEF, which consists of a spatial path, a frequency path and several spatial-frequency fusion modules. Specifically, the spatial branch utilizes DenseNet \cite{huang2017densely} to extract local semantic features from the under-/over-exposed images, facilitating further information integration. Frequency path employs deep Fourier transformation to guide the over-/under-exposed images, focusing on the global attributions over the integration of amplitude and phase components. To further enhance complement information in the frequency and spatial path, the Spatial-Frequency Fusion Module (SFFM) is designed to enable sufficient information interaction between the spatial and frequency domain, which is composed of spatial and channel attention mechanisms to exploit and suppress redundancy information. 

In summary, the main contributions of our work are as follows:
\begin{enumerate}[\textbullet]
\item We propose a novel approach for multi-exposure image fusion by leveraging the deep Fourier transformation. To the best of our knowledge, this is the first work to introduce the spatial and frequency domain information jointly in the MEF network;
\item We design multi-exposure image fusion via the spatial-frequency integration framework (MEF-SFI) to jointly utilize spatial and frequency domain information generating fusion results that maintain appropriate exposure levels. Multiple Spatial-Frequency Fusion Module is cascaded to enable global and local information interaction sufficiently;
\item A Fourier-based loss function is introduced to constrain the global frequency difference between the fusion results and over-/under-exposed image pairs;
\item Extensive experiments demonstrate that our method achieves informativeness and exposure-consistent fusion results in the PAQ-MEF dataset.
\end{enumerate}

\section{Related Work}
Existing MEF methods can be categorized into traditional methods and deep learning-based methods. 

\subsection{Traditional multi-exposure image fusion}
Traditional MEF methods consist of spatial domain and transform domain-based algorithms \cite{zhang2021benchmarking}, which rely on hand-crafted features and fusion strategy. 

The MEF methods based on the spatial domain could be further classified into pixel-based methods, patch-based methods and optimization-based methods. Pixel-based approaches fused image sequences at the pixel level using appropriate fusion rule \cite{li2012fast, bruce2014expoblend}. Kinoshita \cite{kinoshita2019scene} proposed a scene segmentation-based luminance adjustment method to generate informative and perceptually appealing images even when the input images are unclear, which consists of two segmentation approaches separating scenes according to the luminance value and distribution, respectively. Ulucan et al. \cite{ulucan2021multi} extracted fusion maps through the linear embeddings of image pixel/patch spaces and then utilized watershed masking procedure to adjust fusion maps for refining informative information. Patch-based methods segment images into multiple local regions and apply different information measurement and fusion rules at the same position among different images in the sequence. Ma et al.\cite{ma2017robust} decomposed images into signal strength, signal structure and mean intensity three components, and then processed them on the basis of patch strength, exposedness and structural consistency measures. Li et al. \cite{li2021detail} incorporated edge-preserving and bell curve functions with previous patch-decomposition-based methods to preserve detail information in both bright and dark regions. Additionally, \cite{li2012detail, shen2011generalized} are typical optimization-based methods.

However, the above spatial domain-based methods might be prone to artifacts at the junction of different image areas. The transform domain-based methods could be categorized into multi-scale decomposition-based methods, gradient domain-based methods, sparse representation-based methods and other transform-based methods, and these methods are composed of three stages: image transformation, coefficient fusion and inverse transformation \cite{liu2020multi}. Burt et al. \cite{burt1993enhanced} was an early MEF method applying the directional filtering-based gradient pyramid model. Li et al. \cite{li2013image} decomposed images into the base layer and detail layer for capturing intensity information on the large scale and detail information on the small scale, respectively. Then, a guided filtering-based weighted average technique was utilized to integrate information from the dual layer. Gu et al. \cite{gu2012gradient} and Paul et al. \cite{paul2016multi} designed gradient-based methods, where gradients were estimated by the structure tensor and maximum amplitude gradient selection. Spatial representation-based methods combine different images via the sparse weighted coefficients chosen from the over-complete dictionary and non-zero coefficients represent complementary information of images. Wang et al. \cite{wang2014exposure} trained dictionary by using appropriating K-SVD and combined coefficients with "frequency of atoms usage" fusion strategy. 


\subsection{Deep learning-based multi-exposure image fusion}
Recently, deep learning has achieved great progress in different kinds of high-level \cite{gao2022object, xu2020cbfnet} and low-level \cite{zhao2021didfuse, li2021different, huang2022exposure, li2023lrrnet, jiang2022target, ma2019fusiongan} computer vision tasks. The advantage of the deep learning-based method in image fusion is capable of adaptively learning a reasonable feature extraction and fusion strategy, thereby overcoming drawbacks of traditional methods \cite{zhang2021image}. For the MEF, DL-based methods are able to generate smooth transitions in regions of varying exposure levels and can basically be divided into AE-based, CNN-based and GAN-based methods \cite{zhang2021image}. Kalantari et al. \cite{ram2017deepfuse} first attempted to implement MEF with a supervised neural network framework named DeepFuse and optimized the network via MEF-SSIM \cite{ma2015perceptual}. DeepFuse conducted luminance fusion on the Y channel of the YCbCr image and chrominance fusion on the Cb and Cr channels. Since then, Xu et al.  \cite{xu2020u2fusion} applied DenseNet \cite{huang2017densely} as the fusion network and proposed a novelty adaptive complementary information measurement to determine the information retention in the loss function. Deng et al. \cite{deng2021deep} achieved super-resolution (SR) and MEF simultaneously to generate a high-resolution and high-dynamic range image via the fully convolutional network. Liu et al. \cite{liu2023holoco} explored intrinsic information of source LDR images and the reference HDR image through contrastive learning, which treated source LDR image, reference HDR image and the fusion HDR image as the negative, positive and anchor, respectively. Han et al. \cite{han2022multi} not only integrated the luminance component on the Y channel, but also implemented color mapping/correction to adjust the final fusion appearance closing to the human visual realism.

\cite{xu2020mef, yin2021two, liu2022attention, yang2021ganfuse} are representative GAN-based MEF methods. Specifically, Xu et al. \cite{xu2020mef} first employed an end-to-end GAN-based architecture to MEF, in which the generator is trained to obtain the visually appealing fusion image, and the discriminator is trained to distinguish the fusion result from the groundtruth (GT). Yang et al. \cite{yang2021ganfuse} applied dual discriminators between the fusion result and input image pairs. Indeed, GAN-Fuse trained the network via an unsupervised manner, i.e. constraining the similarity between the fusion image and input image pairs rather than GT. Liu et al. \cite{liu2022attention} proposed an attention-guided global-local adversarial learning network, in which the global-local discriminator distinguished fusion results from GT at both image- and patch-level to balance the pixel intensity distribution and texture detail. 

Besides, Ma et al. \cite{ma2019deep} proposed using the fully convolutional network to predict weight maps for the fusion of the input sequence with arbitrary spatial resolution and image number. Jiang et al. \cite{jiang2023meflut} introduced 1D lookup table (LUT) in the MEF to acquire high-quality and high-speed fusion results. It first encoded fusion weight maps into a 1D LUT, and then obtained fusion weights of images in the input sequence by querying the 1D LUT directly, which could boost the fusion efficiency compared with previous methods. Qu et al. \cite{qu2022transmef} and SwinFusion \cite{ma2022swinfusion} compensated for the disadvantage of CNN-based methods in integrating long dependencies via the Transformer \cite{vaswani2017attention}.

\subsection{Fourier Transform-based vision methods}
Recently, there has been a growing interest in applying frequency domain-based deep learning methods to several tasks. For instance, Zhou et al. \cite{zhou2022spatial} proposed the first pan-sharping method attempting to exploit both frequency and spatial domains, which devised spatial-frequency information extraction and integration networks to learn complementary representations. Huang et al. \cite{huang2022deep} and Li et al. \cite{li2022embedding} embedded Fourier transform into the network to facilitate frequency properties, where the amplitude component represents most lightness information and the phase component contains structure information. Yu et al. \cite{yu2022frequency} addressed the image dehazing task via the frequency and spatial dual guidance, integrating global and local features within a framework. Wang et al. \cite{wang2023spatial} devised a spatial-frequency mutual network for face super-resolution to capture both global and dependency. Yu et al. \cite{yu2022deep} devised a reliable upsampling method in the Fourier domain, resolving the problem that the Fourier domain did not share the same local texture similarity and scale-invariant property as the spatial domain. Qu et al. \cite{qu2023aim, qu2023rethinking} proposed two frequency domain-based methods, namely AIM-MEF and FCMEF, to address MEF from different perspectives. AIM-MEF exploited adaptively mining the hard-to-learn information in the frequency domain. FCMEF developed a multi-exposure image dataset synthesis strategy based on the Fourier domain, which could convert natural images with normal exposure to images with different exposure levels. To the best of our knowledge, we are the first to introduce the Fourier domain-based neural network architecture and loss function in the MEF. It is worth noting that the frequency domain-based MEF methods mentioned above are trying to mine hard-to-learn information and synthesize multi-exposure image datasets by incorporating frequency information, rather than directly embedding Fourier transform into the neural network.

\section{Methods}

In this section, we provide details of our network and loss function, including both spatial and frequency-based path and loss function items. Specifically, we use a dual path to extract spatial and frequency features of over- and under-exposed features and integrate these features via an attention block to retain the most informative information.

\subsection{Motivation}
\label{section:motivation}
The goal of our work is to integrate complementary illumination information from images with different exposure levels and generate a single image with comparable visual effects both in global and local areas. Previous works mainly employed CNN to fulfill feature extraction, fusion and image reconstruction. However, due to the limitation of the receptive field, existing CNN-based methods are particularly adept at modeling local features within an image, whereas struggle to capture long-range dependency \cite{yu2022frequency}. Models such as \cite{zhao2023cddfuse, ma2022swinfusion, qu2022transmef} have applied transformer to leverage long-range attention to integrate complementary information. However, transformer-based models are usually computationally heavy, which is hard to deploy on resource-limited devices \cite{liu2023survey}. According to the \cite{frigo1998fftw, chi2019fast}, features learned via Fourier transform can acquire the image-wide receptive field to cover the entire image, which facilitates the learning of global illumination distribution.

To this end, we propose a novel multi-exposure image fusion method called the Spatial-Frequency Interaction framework (MEF-SFI) to effectively leverage the global contextual information modeling ability of the frequency domain, which can be seen as a complement to the representation of local information in the spatial domain. Given a grayscale image $ I \in \mathbb R^{H \times W} $, the Discrete Fourier Transform (DFT) of $ I $ is as follow:
\begin{equation} \label{fourier_transform}
\mathcal F \left(x\right)\left(u, v\right) = \frac{1}{\sqrt{HW}} \sum_{h=0}^{H-1}{\sum_{w=0}^{W-1}{I \left(h,w \right)e^{-j2\pi \left(\frac{h}{H}u+\frac{w}{W}v \right)}}}
\end{equation}

\noindent where $ \mathcal F $ denotes the result of DFT, and $ \mathcal F^{-1} $ denotes the inverse DFT. Additionally, the amplitude component $ \mathcal A \big(\mathcal F \left(x\right)\left(u, v\right) \big) $ and phase component $ \mathcal P \big(\mathcal \mathcal F \left(x\right)\left(u, v\right) \big) $ are expressed as:
\begin{equation}
\mathcal A \big(\mathcal \mathcal F \left(x\right)\left(u, v\right) \big) = \sqrt{R^2 \big(\mathcal F \left(x\right)\left(u, v\right)\big) + I^2 \big(\mathcal F \left(x\right)\left(u, v\right)\big)}
\end{equation}
\begin{equation}
\mathcal P \big(\mathcal F \left(x\right)\left(u, v\right) \big) = arctan \left[  \frac{I \big(\mathcal \mathcal \mathcal F \left(x\right)\left(u, v\right) \big)}{ R \big(\mathcal \mathcal \mathcal F \left(x\right)\left(u, v\right) \big)} \right]
\end{equation}

\noindent where $ R\left(x\right) $ and $ I\left(x\right) $ are the real and imaginary parts of the $ F (x)(u, v) $, respectively.

 \cite{oppenheim1981importance, skarbnik2009importance} showed that the brightness contrast information of the image can be expressed by the amplitude spectrum $ \mathcal A $, meanwhile the phase spectrum $ \mathcal P $ is able to represent the structure information of the image \cite{xu2021fourier, yang2020fda}. As is shown in Fig. \ref{fig:analysis}, we revisit the frequency properties of images with different exposure levels via the Discrete Fourier Transform, and two conclusions are arrived at in our observation, which are: 
\begin{enumerate}[1)]
\item After swapping the amplitude component and phase component of over- and under-exposed images, the pseudo-under image (with the amplitude of under-exposed and the phase of over-exposed) exhibits a similar brightness to the original under-exposed image, and meanwhile the brightness of the pseudo-over image (with the amplitude of over-exposed and the phase of under-exposed) is close to the original over-exposed image;
\item The fusion-over image (with the sum amplitude of over- and under-exposed images and the phase of the over-exposed image) and fusion-under image (with the sum amplitude of over- and under-exposed images and the phase of the under-exposed image) exhibit a similar brightness level that falls within the range of the over- and under-exposed image.
\end{enumerate}

The above observations indicate that images with different exposure levels share a similar structure, which corresponds to the phase spectrum. By integrating the amplitude spectrum using a comparable representation, we can achieve a visually pleasing fusion image.

\begin{figure*}[!t]
\centering
\includegraphics[width=5.5in]{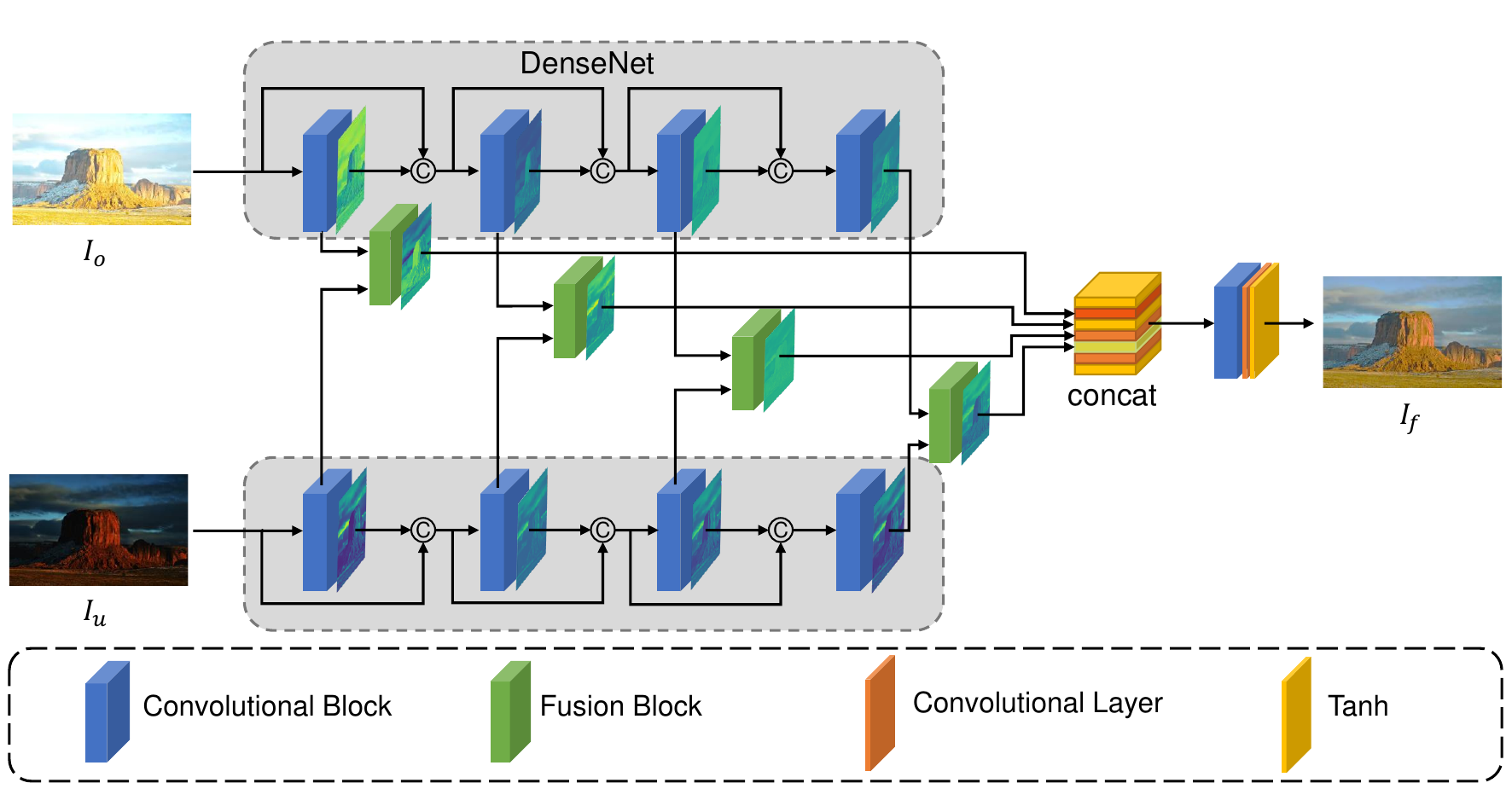}
\caption{
The architecture of the proposed multi-exposure image fusion network. Our network is composed of two feature extraction paths, and each of them corresponds to an over-/under-exposed image as the input. Each path has four convolutional blocks and Spatial-Frequency Fusion Module. After obtaining the fused representations from four fusion blocks, we adopt the concatenation operator and a single convolutional block to generate the final fused image. 
}
\label{fig:architecture}
\end{figure*}

\subsection{Overview}
\label{section:overview}
The architecture of our proposed method is a dual path fully convolutional neural network, as is shown in Fig. \ref{fig:architecture}. Given a three-channel RGB image $  I_o, I_u \in \mathbb R^{H \times W \times 3} $, we first convert the color space of source images from RGB to YCbCr, and only adopt the proposed fusion method on the Y channel. We use the Y channel of the over-exposed image $ I_o^Y \in \mathbb R^{H \times W \times 1} $ and under-exposed image  $ I_u^Y \in \mathbb R^{H \times W \times 1} $ as inputs for the network, where $ H $ and $ W $ are the height and width of images, and then two feature extraction path is applied for each of the input image. Feature extraction path adopts several $ 3 \times 3 $ convolutional layers with PReLU activation function as the basic feature extraction unit, and similar to some previous methods \cite{xu2020u2fusion, jiang2022target}, we refer to the dense connection of the DenseNet \cite{huang2017densely} to strengthen the connection between different layers in each path. Specifically, our dual feature extraction path consists of four convolutional blocks, each of which corresponds to the $i$-th Spatial-Frequency Fusion Module (SFFM) to integrate features $F_o^i$ and $F_u^i$ both in the spatial and frequency domain, generating a fused representation $F_{f}^i$. 

The architecture of our SFFM is shown in Fig.~\ref{fig:fusionblock}. We design a frequency-global and spatial-local branch to fuse features extracted by the feature extraction path, and the detailed information of SFFM will be discussed in Section \ref{section:sffm}. Fusion features $F_{f}^i$ generated by these fusion blocks are concatenated and apply $ 1 \times 1 $ convolution operation and tanh activation function to obtain the output $I_f^Y \in \mathbb R^{H \times W \times 1}$ of the entire network. For the Cb and Cr channels, we simply fuse them via the weighted average method as follows:
\begin{equation}
C_f = \frac{\big( C_1\left( \left| C_1-\tau \right| \right) + C_2\left( \left| C_2-\tau \right| \right)\big)}{\left( \left| C_1-\tau \right| + \left| C_2-\tau \right| \right)}
\end{equation}

\noindent where $C_1$ and $C_2$ are the Cb or Cr channel values of the input image pair, and $\tau$ is set to 128 in our work. $C_f$ is the fused Cb or Cr channel. Finally, the fused Y channel is concatenated with fused Cb and Cr channel, and converted back to the RGB color space to obtain the final high-dynamic image $I_f\in \mathbb R^{H \times W \times 3}$.

Our loss function consists of two parts, which are spatial loss and frequency loss function, respectively. The function of the spatial loss is to impose pixel-level constraints on input image pairs and the fused result, which is the weighted sum of $ MSSIM(I_{o/u}^Y, I_f^Y) $ and $ \left \|I_{o/u}^Y - I_f^Y\right \|_2^2 $. Additionally, the frequency loss is to impose a global constraint between input image pairs and the fused result in the Fourier domain, and its terms are the $ \left \|\mathcal A(I_{o/u}^Y) - \mathcal A(I_f^Y)\right \|_2^2 $ and $ \left \|\mathcal P(I_{o/u}^Y) - \mathcal P(I_f^Y)\right \|_2^2 $.

\subsection{Spatial-frequency fusion module}
\label{section:sffm}
\begin{figure}[!t]
\centering
\includegraphics[width=5.5in]{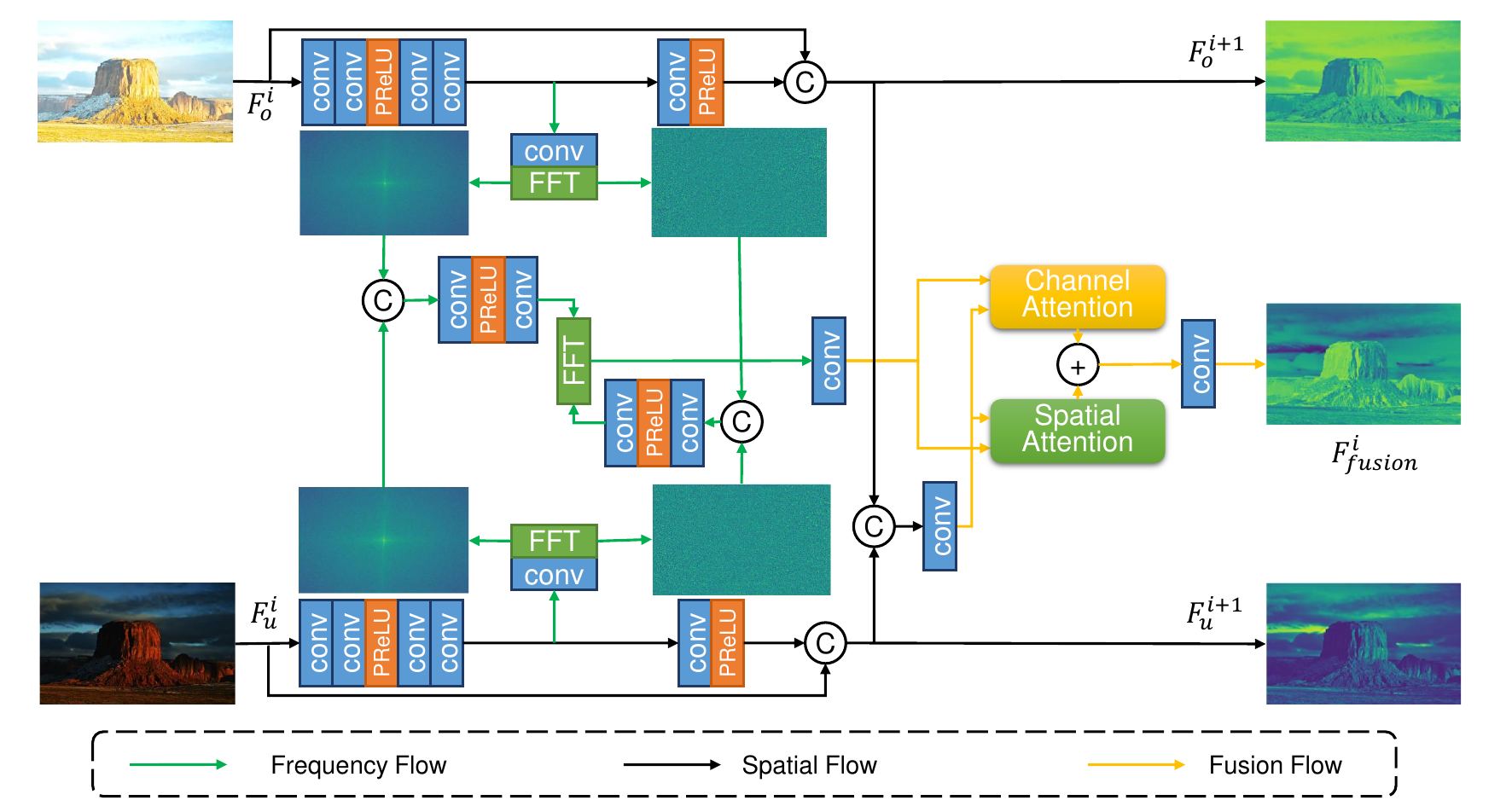}
\caption{
Structure of the Spatial-Frequency Fusion Module. The SFFM consists of a spatial fusion branch, a frequency fusion branch and an attention block. The attention block is used to measure and integrate representations from the spatial and frequency fusion branches.
}
\label{fig:fusionblock}
\end{figure}

The architecture of our spatial-frequency fusion module is shown in Fig.~\ref{fig:fusionblock}, which involves three basic elements: (1) spatial fusion branch for the local information fusion; (2) frequency fusion branch for global illumination information fusion; (3) attention-based fusion block is designed for complementary information measurement and integration. We denote the $F_o^i$ and $F_u^i$ as the $i$-th input features of the over-exposed and under-exposed path, respectively. The output of the SFFM is denoted as the $ F_{f}^i $, which would be concatenated with the output of other SFFM to generate the final fused image $I_f^Y$.

Specifically, the spatial branch directly concatenates the output features $F_o^{i+1}$ and $F_u^{i+1}$ from the basic feature extraction unit as the fused spatial representations, and applies a $ 3 \times 3 $ convolutional layer to form the final fused representation $ F_{spa}^i $ in the spatial domain, which is expressed as follows:
\begin{equation}
 F_{spa}^i = f^{3 \times 3} \left( concat \left[ F_o^{i+1}, F_u^{i+1} \right] \right)
\end{equation}

Additionally, each spatial path also conveys over-/under-exposed features $ F_o^{i+1} $ and $ F_u^{i+1} $ to the next dense block in the dual feature extraction path.

The frequency fusion branch aims to effectively extract and fuse global illumination information via the Deep Fourier Transform. Although the transformer-based methods \cite{qu2022transmef, ma2022swinfusion} are proficient in cross-domain long-range learning, the computational complexity limits their application. As we discussed in section \ref{section:motivation}, the amplitude component $\mathcal A$ in the Fourier domain represents the global illumination information, and the brightness level in different images could be balanced via the combination of $ \mathcal A_o $ and $ \mathcal A_u $. Therefore, we design a frequency branch for capturing and integrating global illumination representations. In Fig.~\ref{fig:fusionblock}, the intermediate features of $ F_o^i $ and $ F_u^i $ of the $i$-th dense block are inputs of the frequency branch, and transform them to frequency domain through the DFT $ \mathcal F (x) $, as Eq. (\ref{fourier_transform}) shows. Then, the amplitude and phase components could be obtained as follows:
\begin{equation}
\begin{aligned}
\mathcal A \left( F_o^i \right), \mathcal P \left( F_o^i \right) &= \mathcal F \left( F_o^i \right) \\
\mathcal A \left( F_u^i \right), \mathcal P \left( F_u^i \right) &= \mathcal F \left( F_u^i \right)
\end{aligned}
\end{equation}

\noindent where $ \mathcal A( \cdot ) $ and $ \mathcal P( \cdot ) $ are the amplitude and phase components of the input features, respectively. After obtaining the amplitude and phase components, we fuse them via the concatenate operation to aggregate the complementary information in the frequency domain from the over-exposed and under-exposed images. Then, we adopt $ 1 \times 1 $ convolution with PReLU to the fused amplitude component $ \mathcal A_f $ and phase component $\mathcal P_f $, which is expressed as follows:
\begin{equation}
\begin{aligned}
\mathcal A_f \left( F_o^i, F_u^i \right)&=CB \left[ concat \big( \mathcal A \left( F_u^i \right), \mathcal A \left( F_o^i \right) \big) \right] \\
\mathcal P_f \left( F_o^i, F_u^i \right)&=CB \left[ concat \big( \mathcal P \left( F_u^i \right), \mathcal P \left( F_o^i \right) \big) \right]
\end{aligned}
\end{equation}

\noindent where $ CB $ denotes the $ 1 \times 1 $ convolutions with PReLU. Lastly, the fusion result of the frequency branch is transformed back to the spatial domain by applying the inverse DFT to the fused amplitude component $ \mathcal A_f \left( F_o^i, F_u^i \right) $ and phase component $ \mathcal P_f \left( F_o^i, F_u^i \right) $ as follows:
\begin{equation}
F_{fre}^i = f^{1 \times 1} \Big( \mathcal F^{-1} \big( \mathcal A_f \left( F_o^i, F_u^i \right), \mathcal P_f \left( F_o^i, F_u^i \right) \big) \Big)
\end{equation}

As we discussed before, the fused frequency representation $ F_{fre} $ captures the global illumination information, whereas lacks detail dependency modeling ability in local regions. Meanwhile, the fused spatial representation $ F_{spa} $ is capable of attending local information to complement texture details. To further enforce the complementary information measurement and integration from the dual domain, we introduce the spatial and channel attention mechanism to the fused spatial representation and frequency representation extracting the most distinguished and informative information \cite{woo2018cbam}. 

Spatial attention mechanism is utilized to exploit the inter-spatial relationship between the fused spatial representations $F_{spa}^i$ and frequency representations $F_{fre}^i$. We refer to the Split-Fuse-Select operator in the SKNet \cite{li2019selective} to aggregate both global and local representations and generate attention maps $S_{spa}^i$ and $S_{fre}^i$ for each input feature to select the most informative spatial area. The generation of the spatial attention map could be described as follows:
\begin{equation}
S_{spa}^i, S_{fre}^i = \sigma \bigg( f^{3 \times 3} \Big( concat \left( F_{spa}^i, F_{fre}^i \right) \Big) \bigg)
\end{equation}

\noindent where $concat$ denotes the concatenation operator, and $f^{3 \times 3}$ is a downsample convolutional block with $3 \times 3$ kernel size. $\sigma$ is the sigmoid function.

Channel attention is designed to suppress redundant information and choose vital information in terms of the channel dimension. Similar to the spatial attention, we first aggerate the features $F_{spa}^i$ and $F_{fre}^i$ from dual branch via the summation operator and squeeze the aggerated features from 2D spatial maps to the 1D channel descriptor via the global average pooling (GAP). Then, we use a fully connected layer to reduce the dimension of the channel and expand it back to channel dimension, similar to the $F_{spa}^i$ / $F_{fre}^i$, which is as follows:
\begin{equation}
C_{spa}^i, C_{fre}^i = \sigma \bigg( MLP \Big( GAP \left( F_{spa}^i + F_{fre}^i \right) \Big) \bigg)
\end{equation}

\noindent where $GAP$ denotes the global average pooling and $MLP$ is a convolutional block consisting of fully connected layers to reduce and expand the number of channels.

\begin{figure*}[!t]
\centering
\includegraphics[width=5.5in]{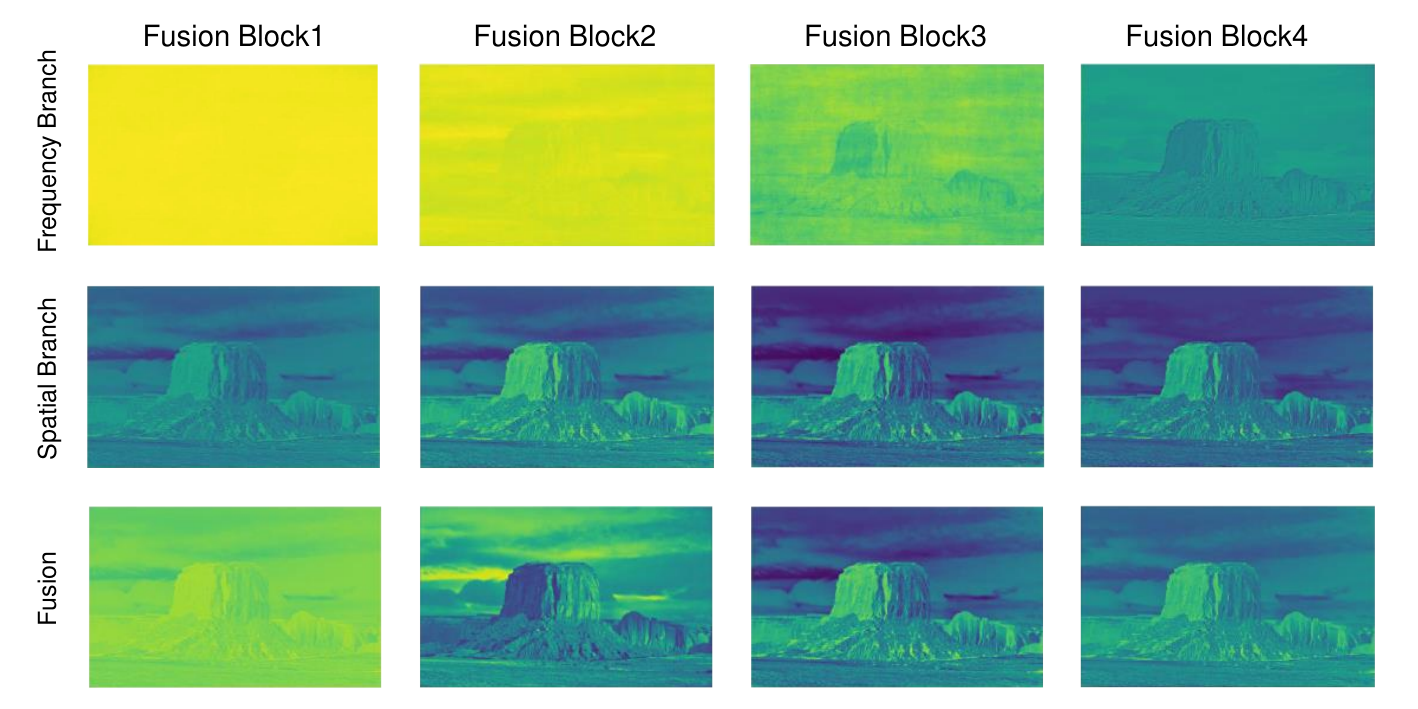}
\caption{
Visualization of feature maps in the different fusion blocks. The visualization results on the top two rows correspond to the frequency fusion branch and the spatial fusion branch, respectively. The last row is feature maps of fusion $F_{fusion}$ after the attention block.
}
\label{fig:feature_map}
\end{figure*}

After generating the spatial and channel attention map, the fused representation $F_{f}^i$ is composed as:
\begin{equation}
F_{f}^i = f^{ 3\times 3 } \bigg( C_{spa}^i \otimes \left( S_{spa}^i \otimes F_{spa}^i \right) + C_{fre}^i \otimes \left( S_{fre}^i \otimes F_{fre}^i \right)  \bigg)
\end{equation}

\noindent where $\otimes$ denotes the element-wise multiplication. The 1D channel attention map is broadcasted over the same 2D dimension as the spatial attention map during the multiplication. The visualization output feature maps of each fusion block are illustrated in Fig. \ref{fig:feature_map}.

\subsection{Loss function}
We propose a spatial-frequency loss to jointly train our network balancing the global/illumination information and local/detail information. Our loss function consists of two parts, which are the spatial part and the frequency part.

The spatial loss aims to constrain the pixel difference between the input image pairs and fusion results, which includes pixel intensity information and detail information. To preserve detail information, we adopt the mean structural similarity index metric (MSSIM) \cite{wang2004image} between the input image pairs and fusion results to describe the perceptual quality of the image. The MSSIM we used in the loss function is averaging SSIM value across all patches at different locations between two images:
\begin{equation}
\label{eq:mssim}
MSSIM\left(X,Y\right) = \frac{1}{J} \sum_{j=1}^{J}{SSIM\left( x_j, y_j \right)}
\end{equation}
 
\noindent where $X$ and $Y$ are full size images, and $x_j$ and $y_j$ are $j$-th  patches in $X$ and $Y$, respectively. SSIM includes luminance, contrast and structure three parts, which can be defined as follows:
\begin{equation}
SSIM\left(x,y\right) = \frac{ \left(2 \mu _x \mu_y + C_1 \right) \left( 2 \sigma_{xy} + C_2 \right) }{\left(2 \mu _x^2 \mu_y^2 + C_1 \right) \left( 2 \sigma_x^2 + \sigma_y^2 + C_2 \right)}
\end{equation}

\noindent where $x$ and $y$ are the image patches of two images. $\mu_x$ and $\mu_y$ are the mean value of the image patches $x$ and $y$, and $\sigma_x$ and $\sigma_y$ are the standard deviation of the image patches $x$ and $y$. $\sigma_{xy}$ is the covariance value between $x$ and $y$, and $C_1$ and $C_2$ are constant values.

Indeed, we utilize L2-norm to constrain the pixel difference between fusion results and input image pairs. Hence, the spatial loss function is calculated as follows: 
\begin{equation}
\begin{aligned}
&\mathcal L_{spa} = \Big( 1-MSSIM\left(I_f^Y, I_o^Y\right) \Big) + \Big( 1-MSSIM\left(I_f^Y, I_u^Y\right)\Big)
\\
&+ \alpha \left( \left \| I_f^Y - I_o^Y \right \|_2^2 + \left \| I_f^Y - I_u^Y \right \|_2^2 \right)
\end{aligned}
\end{equation}

\noindent where $\alpha$ is the parameter controlling the trade-off between the MSSIM term and the L2-norm term. In this work, the value of the $\alpha$ is set to 0.8.

For the frequency loss, we constrain the L2-norm between the amplitude component of fusion results and input image pairs, and also the phase component, which is described as follows:
\begin{equation}
\begin{aligned}
\mathcal L_{amp} \left( F, X, Y \right) &= \left \| \mathcal A \left( F \right) - \mathcal A \left( X \right) \right \|_2^2 + \left \| \mathcal A \left( F \right) - \mathcal A \left( Y \right) \right \|_2^2
\\
\mathcal L_{pha} \left( F, X, Y \right) &= \left \| \mathcal P \left( F \right) - \mathcal P \left( X \right) \right \|_2^2 + \left \| \mathcal P \left( F \right) - \mathcal P \left( Y \right) \right \|_2^2
\end{aligned}
\end{equation}

Therefore, the frequency loss between fusion results and the input image pairs is as follows:
\begin{equation}
\mathcal L_{fre} = \mathcal L_{amp} \left( I_f^Y, I_o^Y, I_u^Y \right) + \mathcal L_{pha} \left( I_f^Y, I_o^Y, I_u^Y \right)
\end{equation}

Finally, our overall loss is the weighted sum of spatial and frequency loss, which can be described as follows:
\begin{equation}
\mathcal L = \mathcal L_{spa} +\gamma \mathcal L_{fre}
\end{equation}

\noindent where $\gamma$ is the balance paremeter between $\mathcal L_{spa}$ and $\mathcal L_{fre}$, and it is set to 0.1 in this paper.

\section{Experiments}
In this section, we first introduce the implementation details of our network, which includes dataset, training parameters, evaluation metrics and state-of-the-art methods for comparison. Then, the qualitative and quantitative results with 8 state-of-the-art methods are presented on the PQA-MEF dataset. Lastly, the ablation study is discussed to evaluate the effectiveness of each part in our framework.

\subsection{Implementation}

\subsubsection{Dataset}

We use PQA-MEF dataset \footnote{https://drive.google.com/drive/folders/1Ik0D2pf93aLOlexevpAE5ftckMTQscZo} from \cite{deng2021deep} to train and validate our model. PQA-MEF is an image-resized version of the SICE dataset \cite{cai2018learning} with the 1/4 scaling factor of the original image size, and select image pairs with extremely bright and dark from the SICE image sequence. Among the dataset, 374 samples are used for training and the other 100 samples are used for evaluation. Each image pair is well aligned.

\subsubsection{Training details}

In the training stage, we randomly crop images to a patch of $256 \times 256$ as the input of the network, and the batch size is set to 80. We use the AdamW optimizer with a learning rate of $1e^{-3}$ and weight decay of $9e^{-3}$ for 600 epochs to optimize the network. We trained our method on the Intel Xeon Gold 6226R CPU and NVIDIA GTX 3090 GPU via the PyTorch.

\subsubsection{Evalutaion metrics}
Since the widely accepted evaluation metrics have not yet been excluded, we select eight object metrics to compare our method with other state-of-the-art methods for performing the effectiveness of our method, including mutual information (MI), feature mutual information (FMI) \cite{haghighat2011non, qu2002information}, gradient-based similarity measurement ($Q^{AB/F}$) \cite{xydeas2000objective}, MEF structural similarity index measure (MEF-SSIM) \cite{ma2015perceptual}, mean structural similarity index metric (MSSIM) \cite{wang2004image}, Yang's metric ($Q_Y$) \cite{li2008novel}, visual information fidelity (VIF) \cite{han2013new} and $Q_{CV}$ \cite{chen2007human}. Among them, MI and FMI are information theory-based methods, which represent the information transferred from the source image, and a larger value means more information is transferred from the source image. FMI is calculated by the feature maps, indicating the feature information transferred from the source image. $Q^{AB/F}$ is an image feature-based metric, indicating the amount of edge information transferred from the source image to the fused image by calculating the weighted sum of the edge strength and orientation values at each location. $Q_Y$, MEF-SSIM and MSSIM measure the structural similarity between two images from different aspects. MEF-SSIM is the most widely accepted measurement metric in MEF, measuring structure and contrast distortion, and its definition is as shown in Eq. (\ref{eq:mssim}).  $Q_Y$ measures the structural information based on the SSIM, which is defined as follows:
\begin{equation}
Q_Y = 
\begin{cases}
\lambda (\omega) SSIM(A,F|\omega) + \left(1-\lambda(\omega)\right) SSIM(B, F|\omega),~ if ~~SSIM(A,B|\omega) \geq 0.75 \\[2ex] 
max\left(SSIM(A,F|\omega), SSIM(B,F|\omega)\right),~ if ~~SSIM(A,B|\omega) < 0.75
\end{cases}
\end{equation}

\noindent where $\omega$ is a local window and $\lambda(\omega)$ is calculated as $\lambda(\omega) = \frac{s(A|\omega)}{s(A|\omega)+s(B|\omega)}$, and $s$ is a local image saliency measurement.

The integrated human perception inspired fusion metrics include VIF and $Q_{CV}$. VIF evaluates the informational fidelity of the fused image, keeping consistency with the human visual system, and its goal is to develop a model that computes the degree of distortion between source and fused images. For the aforementioned metrics, with the exception of $Q_{CV}$, a higher value indicates better performance.

\begin{figure*}[!t]
\centering
\includegraphics[width=5.2in]{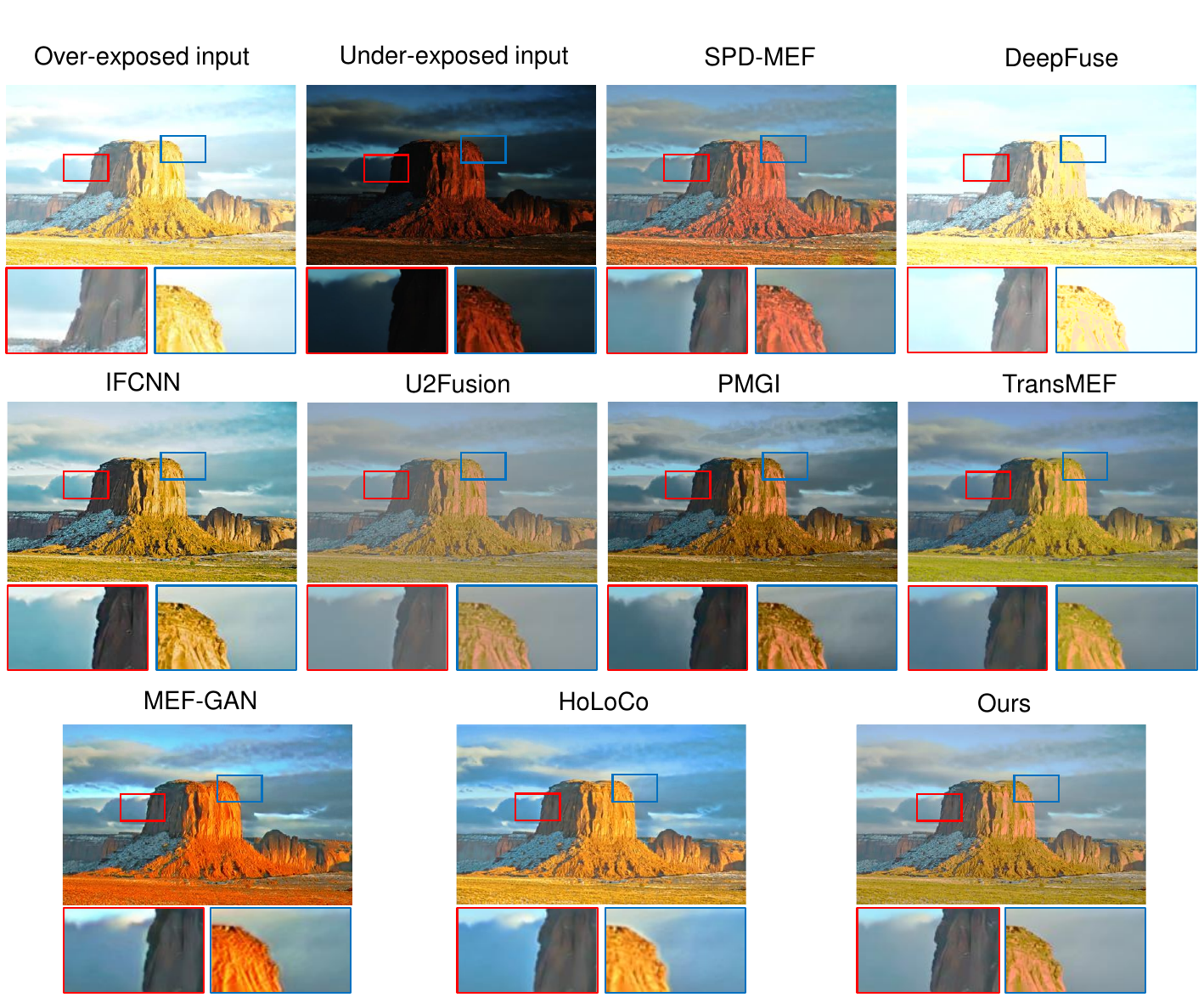}
\caption{
Qualitative results on the image pair 1 on the \emph{PAQ-MEF} test set. The detailed visual results in red and blue boxes are shown at the bottom of each fuesd image.
}
\label{fig:exp_com1}
\end{figure*}

\begin{figure*}[!t]
\centering
\includegraphics[width=5.2in]{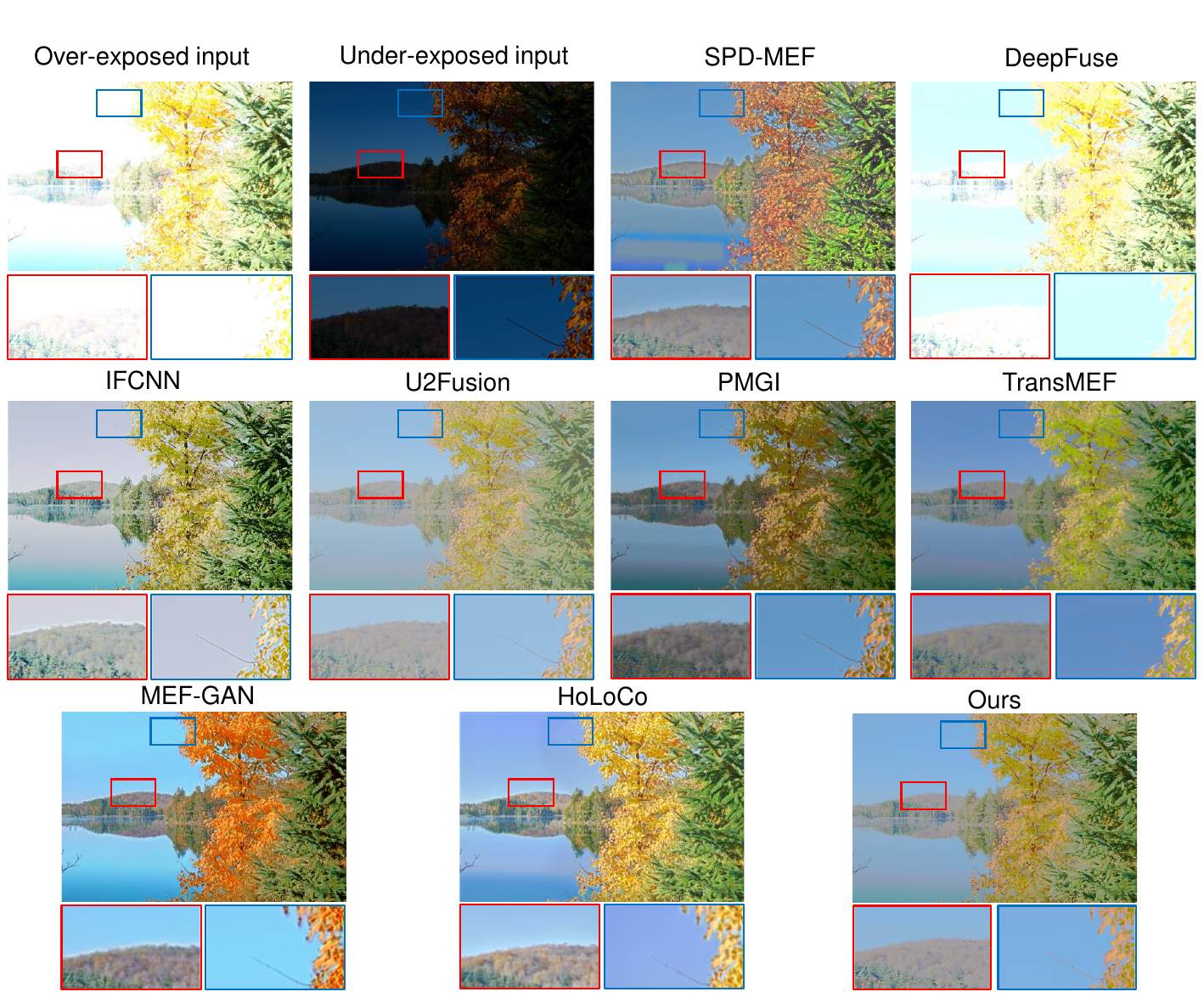}
\caption{
Qualitative results on the image pair 2 on the \emph{PAQ-MEF} test set. The detailed visual results in red and blue boxes are shown at the bottom of each fuesd image.
}
\label{fig:exp_com2}
\end{figure*}

\begin{figure*}[!t]
\centering
\includegraphics[width=5.2in]{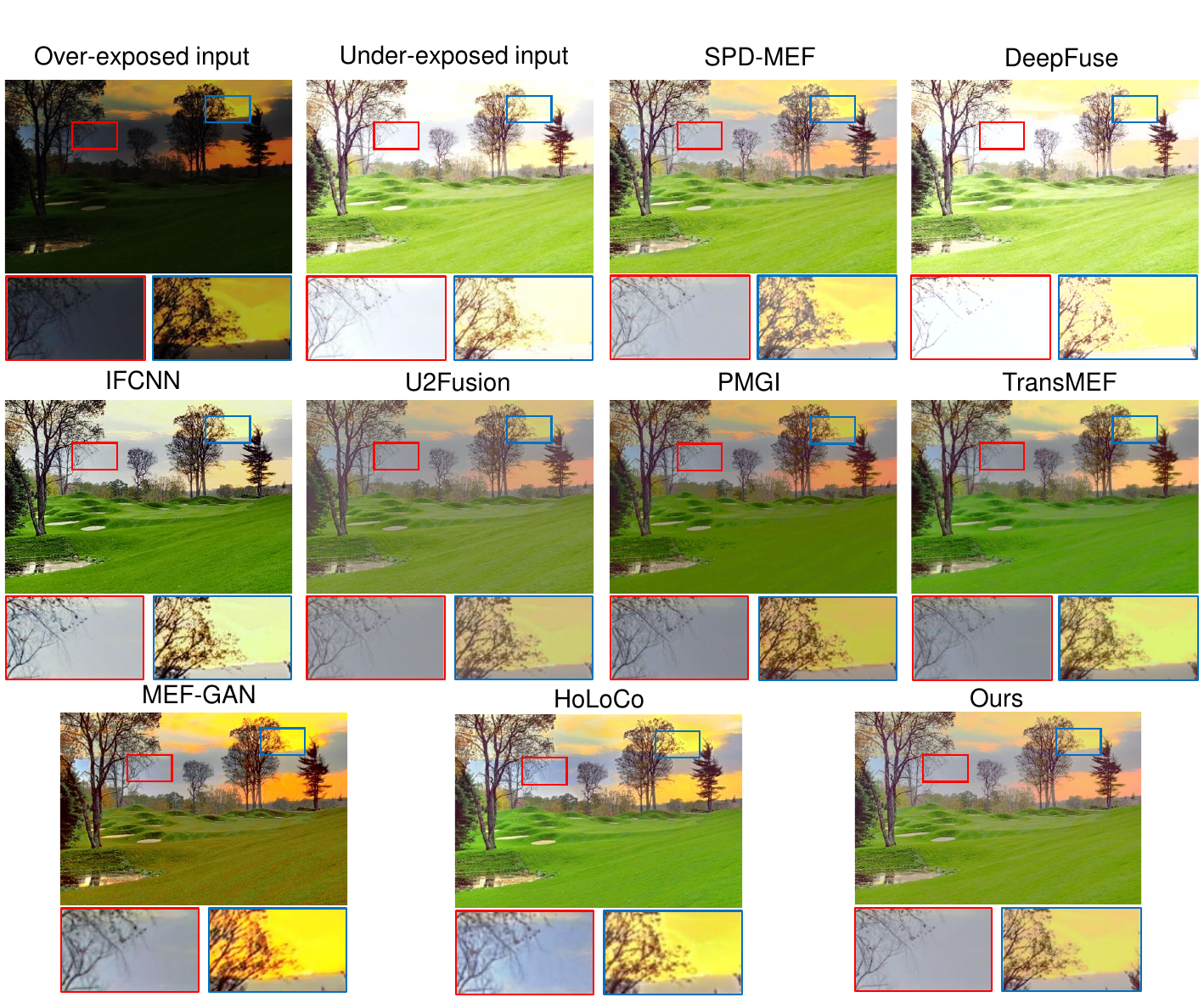}
\caption{
Qualitative results on the image pair 3 on the \emph{PAQ-MEF} test set. The detailed visual results in red and blue boxes are shown at the bottom of each fuesd image.
}
\label{fig:exp_com3}
\end{figure*}

\begin{figure*}[!t]
\centering
\includegraphics[width=5.2in]{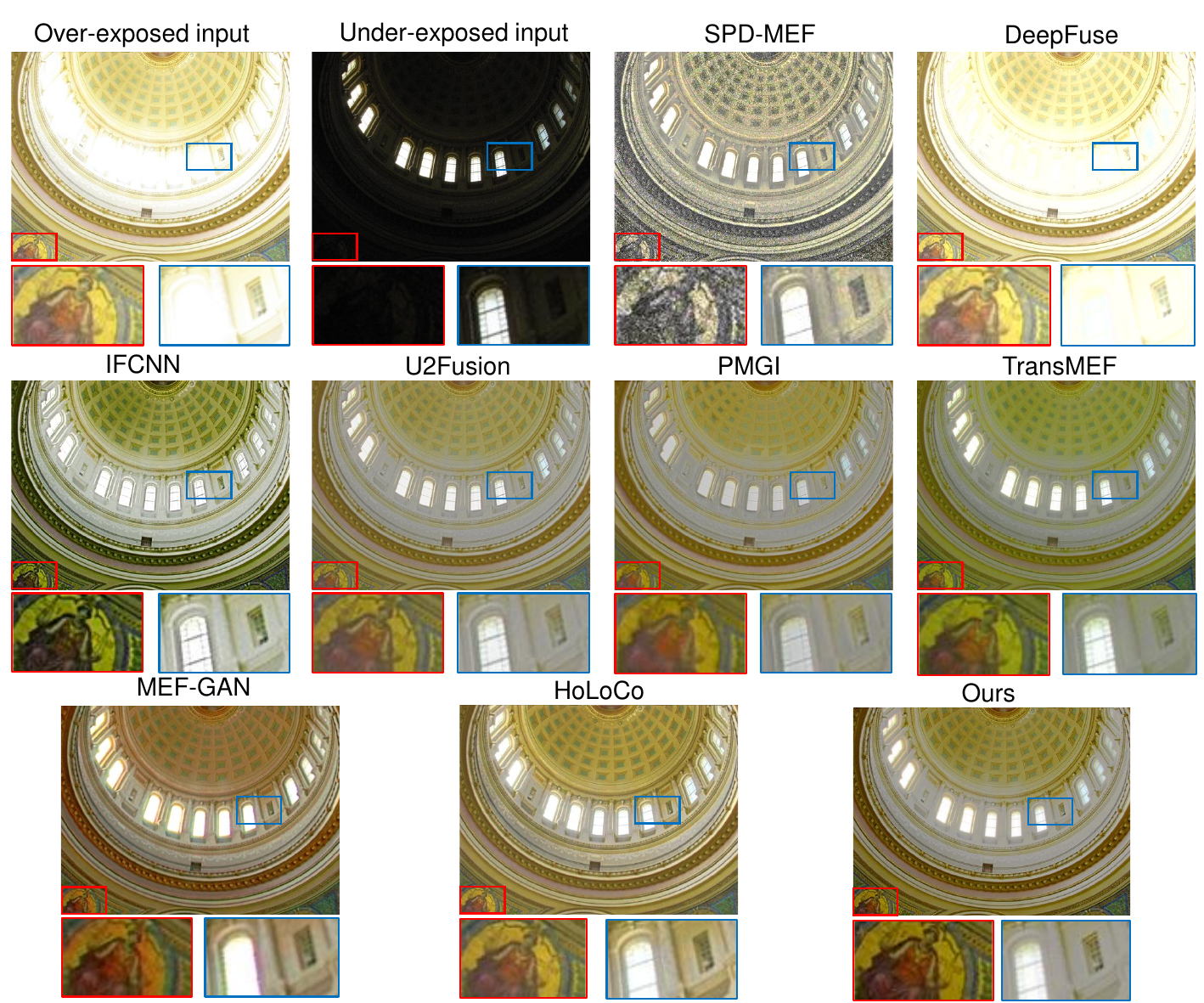}
\caption{
Qualitative results on the image pair 4 on the \emph{PAQ-MEF} test set. The detailed visual results in red and blue boxes are shown at the bottom of each fuesd image.
}
\label{fig:exp_com4}
\end{figure*}

\subsubsection{Comparsion methods}
We compare our methods with 8 state-of-the-art MEF methods, including a traditional method SPD-MEF\cite{ma2017robust} and 7 deep learning methods, which are DeepFuse\cite{zhang2020ifcnn}, IFCNN\cite{zhang2020ifcnn}, U2Fusion\cite{xu2020u2fusion}, PMGI\cite{zhang2020rethinking},  TransMEF\cite{qu2022transmef}, MEF-GAN\cite{xu2020mef}, and HoLoCo \cite{liu2023holoco}. SPD-MEF is a structural patch decomposition method. DeepFuse is the first MEF method to apply deep learning to achieve reliable fusion results. IFCNN, U2Fusoin and PMGI are representative fully convolutional neural network methods. MEF-GAN forms an adversarial relationship and self-attention mechanism driving the network generating real-like fused images. TransMEF applies multi-task learning and transformer to capture global and local information during the fusion. HoLoCo corrects chrominance components besides the illumination component to make fusion results more visually realistic. However, since the motivation of our work is to utilize the global illumination modeling ability of the frequency domain, we only apply our method to the Y channel. Cb/Cr channels are still fused via the weighted sum approach in Section \ref{section:overview}.

\subsection{Reseults on PQA-MEF dataset}

\subsubsection{Qualitative comparison}
We evaluate our method on the PQA-MEF test set and the qualitative comparisons with other methods are performed in Fig.~\ref{fig:exp_com1}, \ref{fig:exp_com2}, \ref{fig:exp_com3}, \ref{fig:exp_com4}. As can be seen from the fusion results, most fusion methods can combine two source images with significant brightness differences into a single image with appropriate brightness, although the quality of fusion results generated by different methods varies significantly. Due to the combination of frequency-global and spatial-local information, our fusion results achieve both visually appealing appearance and promising detail preservation. Additionally, our method successfully avoids generating halo artifacts in the fused image, as indicated by the red boxes in Fig. \ref{fig:exp_com1} and Fig. \ref{fig:exp_com2}, where the corresponding regions in the original input image pair maintain brightness consistency. Traditional MEF method exhibits a lack of robustness in its fusion across various scenarios, like the halo artifact on the water in Fig. \ref{fig:exp_com2} and excessive noise of the indoor construction in Fig. \ref{fig:exp_com4}. Some previous deep learning-based methods like DeepFuse fail to balance different illumination levels. As we can see in Fig.~\ref{fig:exp_com1}-\ref{fig:exp_com4}, the fused images generated by DeepFuse are close to over-expose images, and overexposed regions are unable to recover the detail provided by the under-exposed images. U2Fusion, PMGI and IFCNN are fully convolutional network models used for general image fusion, among which U2Fusion and PMGI design a complementary information measurement in the fusion pipeline to extract vital information from input image pairs effectively. The fused images of these methods preserve details in the over-exposed region, but the excessive intensity contrast causes loss of detail information in the under-exposed region, such as the dark side of the mountain in Fig. \ref{fig:exp_com1}. TransMEF applied transformer module in MEF to compensate the limitation of CNN that can only model local information, and its fusion results are similar with our results. However, as shown in Fig. \ref{fig:exp_com1} and Fig. \ref{fig:exp_com4}, some fused images have color distortion.   HoLoCo implements the color correction module cascaded with the fusion module to ensure the fused images keep consistent with human vision, but the U-shape architecture causes the degradation of detail information in fusion results, such as the branches in Fig. \ref{fig:exp_com3} and painting in Fig. \ref{fig:exp_com4}. Note that our results only apply the proposed method on the Y channel and the color components are fused by the weighted sum. MEF-GAN also fails to protect texture detail as shown in Fig.~\ref{fig:exp_com1}-\ref{fig:exp_com4}. Compared with previous state-of-the-art methods, our method strikes a balance between intensity contrast and detail by incorporating the Fourier-based frequency branch with the simple fully convolutional neural network.

\begin{table}[htbp]
\renewcommand\arraystretch{0.7}
\centering
\caption{Quantitative comparison of our method with nine state-of-the-art models on the \emph{PQA-MEF} dataset. The best performance are shown in \textbf{bold}, and the second and third best performance are shown in \textcolor[rgb]{ 1,  0,  0}{red} and \textcolor[rgb]{ 0,  0,  1}{blue}, respectively.}
\label{exp:comparsion}
\scalebox{0.8}{
\begin{tabular}{lcccccccc}
\toprule
Methods & MI & FMI & $Q^{AB/F}$ & MEF-SSIM & MSSIM & $Q_Y$ & VIF & $Q_{CV}$ ($\downarrow$) \\
\midrule
SPD-MEF & 4.765 & 0.863 & \textcolor[rgb]{1, 0, 0}{0.680} & 0.762 & \textcolor[rgb]{1, 0, 0}{0.939} & \textcolor[rgb]{1, 0, 0}{0.799} & \textcolor[rgb]{1, 0, 0}{1.596} & 296.189 \\
DeepFuse & \textcolor[rgb]{1, 0, 0}{6.084} & \textcolor[rgb]{0, 0, 1}{0.870} & \textcolor[rgb]{0, 0, 1}{0.660} & 0.872 & 0.914 & \textcolor[rgb]{0, 0, 1}{0.769} & 1.268 & 302.561 \\
IFCNN & 4.981 & 0.860 & 0.654 & 0.890 & \textcolor[rgb]{0, 0, 1}{0.937} & 0.621 & 1.398 & \textbf{196.616} \\
U2Fusion & 5.135 & 0.859 & 0.500 & \textcolor[rgb]{1, 0, 0}{0.911} & 0.880 & 0.706 & \textbf{1.790} & 281.052 \\
PMGI & 4.982 & 0.856 & 0.441 & \textcolor[rgb]{0, 0, 1}{0.900} & 0.861 & 0.653 & 1.089 & 351.374 \\
TransMEF & \textcolor[rgb]{0, 0, 1}{5.469} & \textcolor[rgb]{1, 0, 0}{0.871} & 0.538 & 0.889 & 0.910 & 0.685 & 1.321 & 326.460 \\
MEF-GAN & 4.538 & 0.848 & 0.464 & 0.874 & 0.905 & 0.505 & 1.077 & 268.588 \\
HoLoCo & 3.952 & 0.841 & 0.400 & 0.890 & 0.907 & 0.563 & 1.078 & \textcolor[rgb]{0, 0, 1}{261.616} \\
Ours & \textbf{6.229} & \textbf{0.891} & \textbf{0.746} & \textbf{0.969} & \textbf{0.954} & \textbf{0.887} & \textcolor[rgb]{0, 0, 1}{1.502} & \textcolor[rgb]{1, 0, 0}{204.895} \\
\bottomrule
\end{tabular}
}
\end{table}

\subsubsection{Quantitative comparison}
Quantitative comparison with other state-of-the-art methods on the PQA-MEF test set is shown in Table \ref{exp:comparsion}. Our method outperforms all other methods in MEF-SSIM, MS-SSIM, $Q_Y$ and $Q^{AB/F}$metrics, which corresponds to the best detail information preservation in the visualization comparison. Values of MI and FMI beyond all other metrics indicate that our method retains a sufficient amount of intensity information in the original image pair. Our method also ranks among the top three in the other two metrics, with VIF and $Q_{CV}$, which suggest that the proposed method can generate fused images consistent with the human visual system. In summary, the proposed method achieves a balance between preserving the brightness information and the detail information of the source image pair.

\subsubsection{Visualization of feature maps}
The visualization results of feature maps in different fusion blocks are illustrated in Fig. \ref{fig:feature_map}. We select one feature map of the frequency fusion branch, spatial fusion branch and final fusion from each of the four fusion blocks to visualize the effectiveness of the spatial-frequency integration. As we can see in  Fig. \ref{fig:feature_map}, the feature maps of the frequency fusion branch can only capture global illumination representations at the first fusion block, and local detail information is gradually modeled after several interactions with the spatial branch, which is reflected in the clearer feature maps from second to fourth columns. Accordingly, the spatial fusion branch also reflects the learning process from local detail information to global illumination information. We can see that the feature map of the first column of the spatial fusion branch is detail-rich and the brightness level is not reasonable. After several fusion blocks, the last fuson feature map is visually appealing.

\begin{figure*}[!t]
\centering
\includegraphics[width=5.5in]{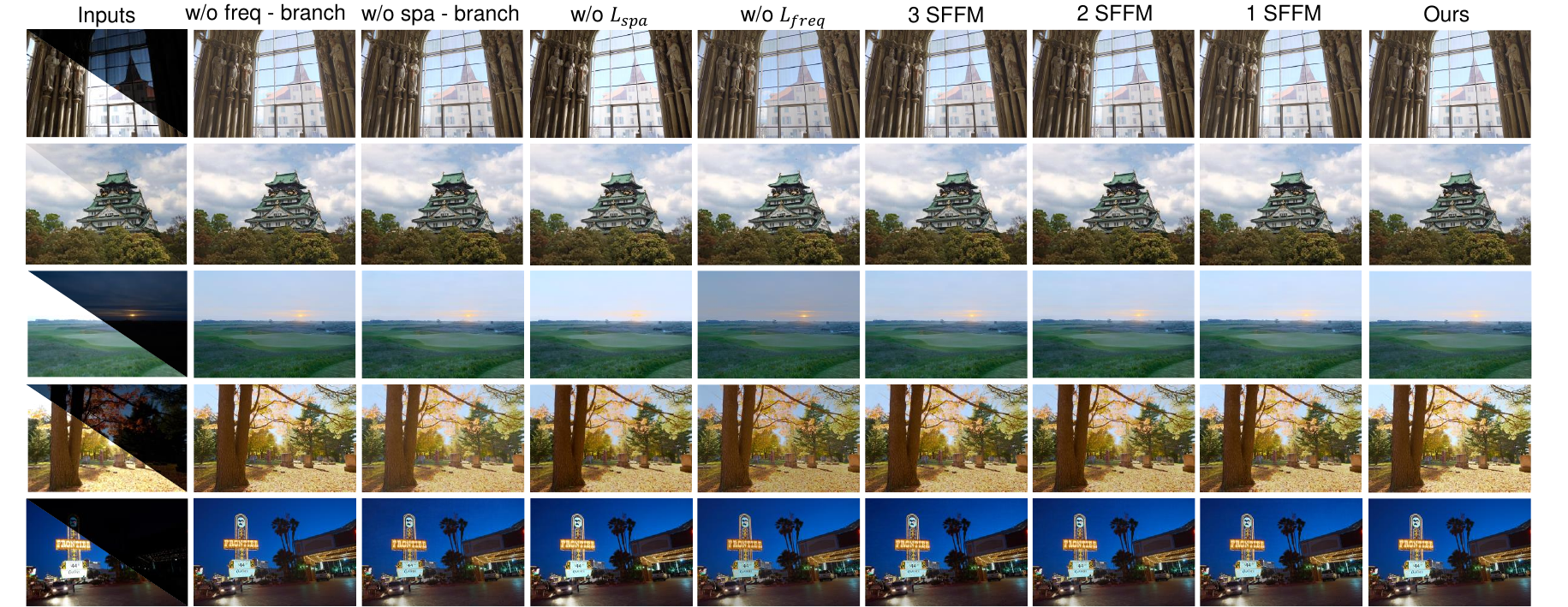}
\caption{
Qualitative results of ablation study on the \emph{PAQ-MEF} dataset. The second and third columns correspond to the ablation studies without the frequency branch and spatial branch, respectively.  The fourth and fifth columns represent ablation studies without the spatial and frequency components of the loss function, respectively. The sixth to eighth columns are ablation studies involving the number of SFFM. Finally, the last column corresponds to the method proposed in this paper.
}
\label{fig:ablation}
\end{figure*}

\subsection{Ablation study}
In this section, we conduct several ablation studies to investigate the effectiveness of the proposed methods. Firstly, to verify the effectiveness of the architecture, we remote the frequency fusion branch and spatial fusion branch in the fusion block to fuse images, respectively. Then, we verify the frequency and spatial part of the loss function by removing the corresponding part. Lastly, the number of SFFM cascaded in the network is also explored for the potential lightweight capabilities of our MEF fusion framework.

\subsubsection{Investigation of network architecture}
In this ablation study, we demonstrate the effectiveness of different architectures in the single SFFM for fusion by removing the frequency branch and spatial branch, respectively. Without incorporating the frequency branch, the value of MSSIM and $Q_Y$ are higher than using the dual branch, and conversely, the value of FMI is the highest when the spatial branch is removed, as shown in Table \ref{tab:ablation}.

\begin{table}[htbp]
\renewcommand\arraystretch{0.7}
  \centering
  \caption{Quantitative results of ablation study on the \emph{PAQ-MEF} dataset.  The best performance are shown in \textbf{bold}, and the second and third best performance are shown in \textcolor[rgb]{ 1,  0,  0}{red} and \textcolor[rgb]{ 0,  0,  1}{blue}, respectively.}
  \scalebox{0.8}{
   \setlength{\tabcolsep}{1.5mm}{
    \begin{tabular}{lcccccccc}
     \toprule
       \multicolumn{1}{c}{Conidition} &  MI & FMI & $Q^{AB/F}$ & MEF-SSIM & MSSIM & $Q_Y$ & VIF & $Q_{CV}$ ($\downarrow$) \\
    \midrule
    $w/o$ freq-branch& \textcolor[rgb]{1, 0, 0}{6.057} & \textcolor[rgb]{1, 0, 0}{0.890} & \textcolor[rgb]{0, 0, 1}{0.742} & 0.970 & \textcolor[rgb]{1, 0, 0}{0.956} & \textbf{0.963} & \textcolor[rgb]{1, 0, 0}{1.444} & \textcolor[rgb]{1, 0, 0}{208.726} \\
    $w/o$ spa-branch& 5.092 & \textbf{0.891} & 0.714 & 0.968 & \textcolor[rgb]{0, 0, 1}{0.954} & 0.716 & 1.402 & 234.465 \\
    \cline{1-1}
    $w/o$ $L_{fre}$ & 5.046 & 0.877 & \textbf{0.751} & \textbf{0.977} & \textbf{0.967} & \textcolor[rgb]{0, 0, 1}{0.862} & \textcolor[rgb]{0, 0, 1}{1.440} & 282.831 \\
    $w/o$ $L_{spa}$ & \textcolor[rgb]{0, 0, 1}{6.046} & 0.883 & 0.741 & 0.965 & 0.944 & 0.752 & \textbf{1.502} & 249.876 \\
    \cmidrule(r){1-1}
    3 SFFM cascaded& 5.183 & 0.887 & 0.723 & \textcolor[rgb]{1, 0, 0}{0.972} & 0.952 & 0.733 & 1.426 & \textcolor[rgb]{0, 0, 1}{222.480} \\
    2 SFFM cascaded& 5.275 & 0.887 & 0.724 & \textcolor[rgb]{0, 0, 1}{0.971} & 0.951 & 0.739 &1.422 & 227.373 \\
    1 SFFM cascaded& 5.147 & \textcolor[rgb]{0, 0, 1}{0.888} & 0.725 & 0.968 & 0.948 & 0.737 & 1.426 & 228.580 \\
    Ours & \textbf{6.229} & \textbf{0.891} & \textcolor[rgb]{1, 0, 0}{0.746} & 0.969 & \textcolor[rgb]{0, 0, 1}{0.954} & \textcolor[rgb]{1, 0, 0}{0.887} & \textbf{1.502} & \textbf{204.895} \\
    \bottomrule
    \end{tabular}%
    }}
  \label{tab:ablation}%
\end{table}%

\subsubsection{Investigation of loss function}
We reveal that without using the frequency part of the loss function to optimize the network, the detail information has been enhanced, as the observation from the largest value of the MEF-SSIM and MSSIM in Table \ref{tab:ablation}. However, information transferred from the input image pair is reduced and global illumination is not comparable, as shown in the first row of the fused image without frequency part in Fig. \ref{fig:ablation} and values of MI and FMI in Table \ref{tab:ablation}. Without considering the spatial part of the loss function, the detail information is diluted as shown in Table \ref{tab:ablation} with values of MEF-SSIM and MSSIM.

\subsubsection{Investigation of number of SFFM}
The fusion results in Fig. \ref{fig:ablation} and Table \ref{tab:ablation} demonstrate that even after removing a specific number of fusion blocks, the method presented in this paper can still obtain comparable fusion results, which opens up the possibility of developing a lightweight version of the fusion network.

\section{Conclusion}
In this study, we propose a multi-exposure image fusion method to comprehensively utilize the global modeling ability of the frequency domain and local modeling of the spatial domain via the Spatial-Frequency Integration Framework. For global modeling, we adopt deep Fourier transform to acquire the amplitude and phase components of the image and combine them via convolution layers. Local modeling is implemented by the fully convolution layers with dense connections. The fused representations from the dual domain are integrated by the attention mechanism. The comparison results with other state-of-the-art methods show that our method is able to achieve visual-appealing fusion results both in qualitative and quantitative experiments, and the ablation study verifies the effectiveness of different parts in our framework.

\section{Declaration of interests}
The authors declare that they have no known competing financial interests or personal relationships that could have appeared to influence the work reported in this paper.

\bibliographystyle{elsarticle-num}

\bibliography{cas-refs}

\end{document}